\documentclass[10pt,twocolumn,letterpaper]{article}

\usepackage{iccv}
\usepackage{times}
\usepackage{epsfig}
\usepackage{graphicx}
\usepackage{amsmath}
\usepackage{amssymb}
\usepackage{booktabs}
\usepackage{subfigure}
\usepackage[pagebackref=true,breaklinks=true,letterpaper=true,colorlinks,bookmarks=false]{hyperref}

\iccvfinalcopy % *** Uncomment this line for the final submission

\begin{document}

%%%%%%%%% TITLE
\title{DeepFaceLab: Integrated, flexible and extensible face-swapping framework}

\author{%
	Ivan Petrov\\
	Freelancer\\
	\texttt{lepersorium@gmail.com} \\
	% examples of more authors
	\and
	Daiheng Gao \\
	Freelancer\\
	\texttt{samuel.gao023@gmail.com} \\
	\and
	Nikolay Chervoniy \\
	Freelancer\\
	\texttt{n.chervonij@gmail.com} \\
	\and
	Kunlin Liu$^{\dagger}$ \\
	USTC\\
	\texttt{lkl6949@mail.ustc.edu.cn} \\
	\and
	Sugasa Marangonda \\
	Freelancer\\
	\texttt{thedeepfakechannel@gmail.com} \\
	\and
	Chris Umé \\
	VFX Chris Ume\\
	\texttt{info@vfxchrisume.com} \\
	\and
	Jian Jiang \\
	008 Tech\\
	\texttt{jiangjian@008tech.com} \\
	\and
	Luis RP \\
	Freelancer\\
	\texttt{luisguans@hotmail.com} \\
	\and
	Sheng Zhang \\
	Freelancer\\
	\texttt{cndeepfakes@gmail.com} \\
	\and
	Pingyu Wu \\
	Freelancer\\
	\texttt{wpydcr@hotmail.com} \\
	\and
	Weiming Zhang \\
	USTC\\
	\texttt{zhangwm@ustc.edu.cn} \\
}

\maketitle

%%%%%%%%% ABSTRACT
\begin{abstract}
Deepfake defense not only requires the research of detection but also requires the efforts of generation methods. However, current deepfake methods suffer the effects of obscure workflow and poor performance. To solve this problem, we present DeepFaceLab, the current dominant deepfake framework for face-swapping. It provides the necessary tools as well as an easy-to-use way to conduct high-quality face-swapping. It also offers a flexible and loose coupling structure for people who need to strengthen their pipeline with other features without writing complicated boilerplate code. We detail the principles that drive the implementation of DeepFaceLab and introduce its pipeline, through which every aspect of the pipeline can be modified painlessly by users to achieve their customization purpose. It is noteworthy that DeepFaceLab could achieve cinema-quality results with high fidelity.
We demonstrate the advantage of our system by comparing our approach with other face-swapping methods.\footnote{For more information, please visit: 	https://github.com/iperov/DeepFaceLab/. (Kunlin Liu is the corresponding author.)}
\end{abstract}

%%%%%%%%% BODY TEXT
\section{Introduction}

\begin{figure*}[htpb]
	\begin{center}
		\includegraphics[width=0.74\linewidth]{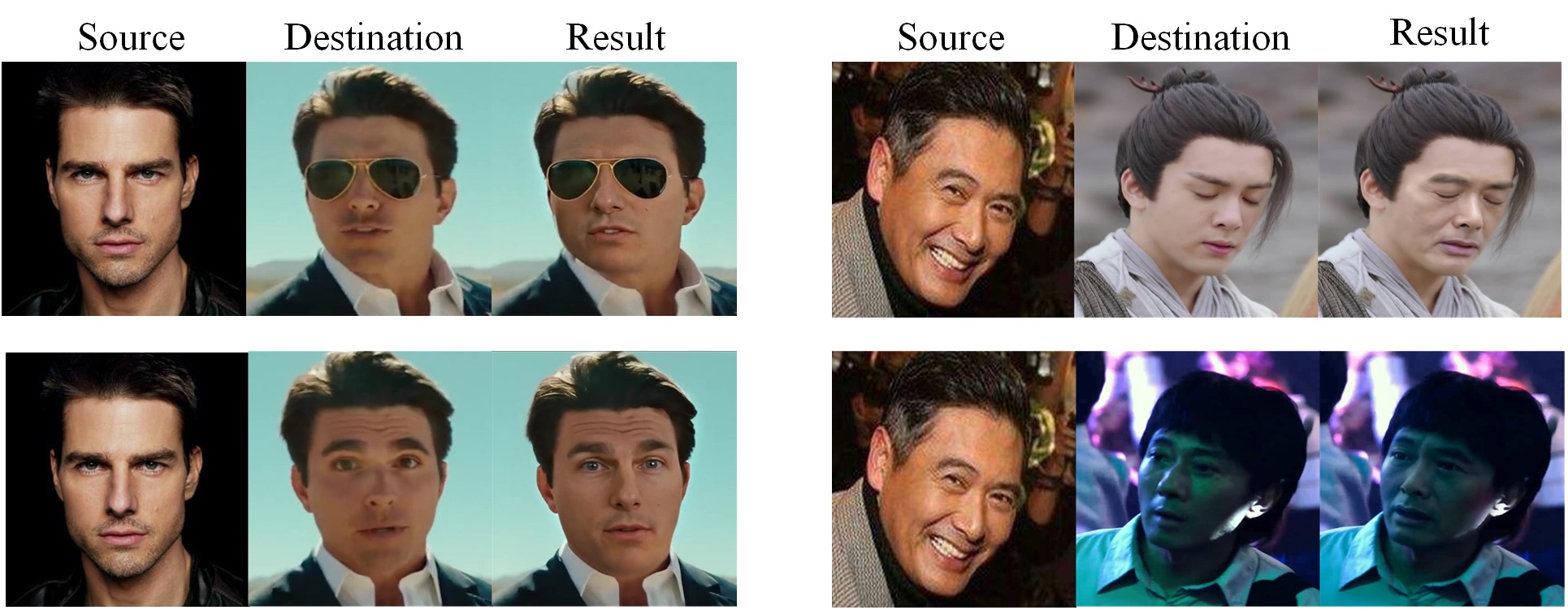}
	\end{center}
	\caption{Face swapping results generated by DeepFaceLab. Left: Source face. Middle: Destination face for replacement. Our results appear on the right, demonstrating that DeepFaceLab could handle occlusion, bad illumination , and side face with high fidelity. 
	%For more information, please visit: 	\url{https://github.com/iperov/DeepFaceLab/}
	}
	\label{fig:long1}
	\label{fig:onecol1}
\end{figure*}
Since deep learning has empowered the realm of computer vision in recent years, manipulating digital images, especially the manipulation of human portrait images, has improved rapidly and achieved photorealistic results in most cases. Face swapping is an eye-catching task in generating fake content by transferring a source face to the destination while maintaining the destination's facial movements and expression deformations. 

The fundamental motivation behind face manipulation techniques is Generative Adversarial Networks (GANs)~\cite{goodfellow2014generative}. More and more faces synthesized by StyleGAN~\cite{karras2019style}, StyleGAN2~\cite{karras2019analyzing} are becoming more and more realistic and entirely indistinguishable for the human vision system. 

Numerous spoof videos synthesized by GAN-based face-swapping methods are published on YouTube and other video websites. Commercial mobile applications such as ZAO$\footnote{https://apps.apple.com/cn/app/id1465199127}$ and FaceApp$\footnote{https://apps.apple.com/gb/app/faceapp-ai-face-editor/id1180884341}$ which allow general netizens to create fake images and videos effortlessly significantly boost the spreading of these swapping techniques, called deepfake.

These content generation and modification technologies
may affect public discourse quality and infringe upon the citizens' right of portrait, especially given that deepfake may be used maliciously as a source of misinformation, manipulation, harassment,
and persuasion. Identifying manipulated media is a technically demanding and rapidly evolving challenge that requires collaborations across the entire tech industry and beyond.

Research on media anti-forgery detection is being invigorated and dedicating growing efforts to forgery face detection. DFDC$\footnote{https://deepfakedetectionchallenge.ai/}$ is a typical example, a million-dollar competition launched by Facebook and Microsoft. Training robust forgery detection models requires high-quality fake data. Data generated by our methods are involved in the DFDC dataset\cite{DFDC}.

However, detection after being attacked is not the unique manner for reducing the malicious influence of deepfake. It is always too late to detect spreading spoofing content. In our perspective, for both academia and the general public, helping netizens know what deepfake is and how a cinema-quality swapped video is generated is much better. 
As the old saying goes:``\textbf{The best defense is a good offense}".
Making general netizens realize the existence of deepfake and strengthening their identification ability for spoof media published in social networks is much more critical than agonizing the fact whether spoof media is true or not.

In 2018, DeepFakes~\cite{Deepfakes} introduced a complete production pipeline in replacing a source person's face with the target person's along with the same facial expression such as eye movement, facial muscle movement. However, the results produced by DeepFakes are poor somehow, so are the results with contemporary Nirkin et al.'s automatic face swapping~\cite{nirkin2018_faceswap}. In order to further awaken people's awareness of 
facial-manipulation videos and provide convenience for forgery detection researchers, we established an open-source deepfake project, DeepfaceLab (DFL for short), which is used to build enormous high-quality face-swapping videos for entertainment and greatly help the development of forgery detection by providing high-quality forgery data.

This paper introduces DeepFaceLab, an integrated open-source system with a clean-state design of the pipeline, achieving photorealistic face-swapping results without painful tuning. DFL has turned out to be very popular with the public. For instance, many artists create DFL-based videos and publish them on their YouTube channels. These videos made by DFL have more than 100 million hits.

The contributions of DeepFaceLab can be summarized as three-folds:

\begin{itemize}
	\item A state-of-the-art framework consists of a maturity pipeline is proposed, aiming to achieve photorealistic face-swapping results.
	
	\item DeepFaceLab open-sourced the code in 2018 and always kept up to the progress in the computer vision area, making a positive contribution for defending deepfake, which has drawn broad attention in the open-source community and VFX areas.
	
	\item A series of high-efficiency components and tools are introduced in DeepFaceLab to build better face-swapping videos.
\end{itemize}

% 2020.02.29 星期六 初版.
\section{Characteristics of DeepFaceLab}

DeepFaceLab’s success stems from weaving previous ideas into a design that balances speed and ease of use and the booming of computer vision in face recognition, alignment, reconstruction, segmentation, etc. There are Four main characteristics behind our implementation:

% keras和tensorflow 2020.03.18
% \paragraph{Leras} Now DeepFaceLab provides a new high-level deep learning framework built on pure TensorFlow~\cite{abadi2016tensorflow}, which aims to bail out the unnecessary restrictions and extra overheads brought by some commonly-used high-level frameworks such as Keras~\cite{chollet2015keras} and plaidML~\cite{Vertex.ai}. We named it Leras, the abbreviation for Lighter Keras. The main advantages of Leras are:
% \begin{itemize}
% 	% 2020.03.18
% 	\item \textbf{Simple and flexible model construction} Leras alleviates the burden of researchers and practitioners by providing Pythonic style to do model work, similar to PyTorch (i.e. defining layers, composing neural models, writing optimizers), but in graph mode (no eager execution).
% 	\item \textbf{Performance focused implementation} With the utilization of Leras instead of Keras, the training time hence reduced by about 10\textasciitilde 20\% on average.
% 	\item \textbf{Fine-granularity tensor management} The motivation for switching to pure Tensorflow is that Keras and plaidML are not flexible enough. In addition, they are largely outdated and do not give full control over how tensors are processed.
% 	% 2019.7.31
% \end{itemize}

\paragraph{Convenience}  DFL strives to make the usage of its pipeline, including data loader and processing, model training, and post-processing, as easy and productive as possible. Unlike other face swapping systems, DFL provides a complete command-line tool with every aspect of the pipeline that could be implemented in the way that users choose. Notably, the complexity inherent and many hand-picked features for fine-grained control such as the canonical face landmark for face alignment should be handled internally and hidden behind DFL. People could achieve the smooth and photorealistic face-swapping results without the need for hand-picked features if they follow the settings of the workflow, but only with the need of two videos: the source video ($\rm src$) and the destination video ($\rm dst$) without the requirement to pair the same facial expression between $\rm src$ and $\rm dst$. To some extent, DFL could function as a point-and-shoot camera. 

% 2020.04.19 week16 交互convert模式.
% Furthermore, according to many practical feedbacks from DeepFaceLab users, a highly flexible and customized face converter is needed since there are a lot of complexity need to be handle: floodlights, rain, separated by glass, face injuries and many other cases. Hence, interactive mode has been applied in the conversion phase, which relieved the workload for deepfake producers since interactive preview can assist them in observing the effects of all changes they make when changing various options and enabling/disabling various features.
\begin{figure*}[h]
	\begin{center}
		\includegraphics[width=0.8\linewidth]{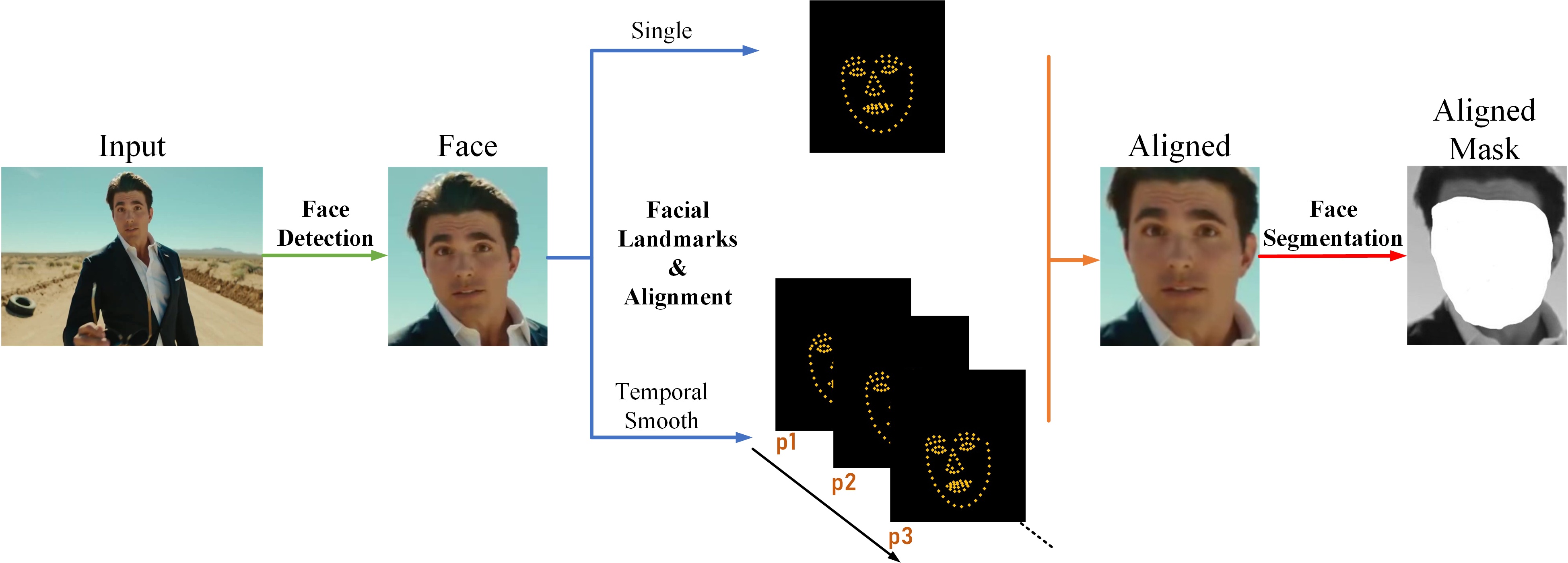}
	\end{center}
	\caption{Overview of extraction phase in DeepFaceLab (DFL for short).}
	\label{fig:long2}
	\label{fig:onecol2}
\end{figure*}
\paragraph{Wide engineering support}
Some practical measures were added to improve the performance: multi-GPU support, half-precision training, usage of pinned CUDA memory to improve throughput, use of multiple threads to accelerate graphics operations and data processing. Even a machine with 2GB VRAM can also conduct a successful face-swapping project.

\paragraph{Extensibility} To strengthen the flexibility of DFL's workflow and attract the interests of the research community, users are free to replace any component of DFL that does not meet their requirements. Most of DFL's modules are designed to be interchangeable.
For instance, people could provide a newer face detector to achieve higher performance in detecting faces with extreme angles or outlying areas. 
%Besides, users could adjust the framework to conduct lip manipulation and head replacement.

\paragraph{Scalability} Having good datasets is essential for the face-swapping task. Generally, the larger the datasets, the better the final results. However, results that are directly extracted from $\rm src$ and $\rm dst$ are always with noises, which significantly harm the final quality. In consideration of the complex situations of input videos, DFL provides a series of measures to clean up datasets. With these measures, DFL has robust scalability and can even support massive scale datasets and conduct cinema-quality face-swapping basing on large datasets.

\section{Pipeline}
\label{section:pipeline}
DeepFaceLab provides a set of workflow which form the flexible pipeline. In DeepFaceLab (DFL for short), we can abstract the pipeline into three phases: extraction, training, and conversion. These three parts are presented sequentially. Besides, it is noteworthy that DFL falls in a typical one-to-one face-swapping paradigm, which means there are only two kinds of data: $\rm src$ and $\rm dst$, the abbreviation for source and destination, are used in the following narrative. 
%Furthermore, unlike prior work, we can generate high-resolution images and generalize to variant input resolutions. 

% 2020.03.06 星期五(week10) 
% 每个部分的作用, 目的, 算法的简单介绍.
\subsection{Extraction}

The extraction phase is the first phase in DFL, aiming to extract a face from $\rm src$ and $\rm dst$ data. This phase consists of many algorithms and processing parts, i.e., face detection, face alignment, and face segmentation. DFL provides many extraction modes (i.e, $\rm half$-$\rm face$, $\rm full$-$\rm face$, $\rm whole$-$\rm face$), which represents the face coverage area of the extraction phase. Generally, we take $\rm full$-$\rm face$ mode by default.
%Plus, as DFL provides many face types (i.e, \verb|half face|, \verb|full face|, \verb|whole face|), which represents the face coverage area of extraction phase. Unless stated otherwise, \verb|full face| is taken by default. 

\paragraph{Face Detection} The first step in extraction phase is to find the target face in the given data: $\rm src$ and $\rm dst$. DFL regards S3FD~\cite{zhang2017s3fd} as its default face detector. S3FD can be replaced with other face detection algorithms painlessly, i.e RetinaFace~\cite{deng2019retinaface}.

% Obviously, you can choose any other face detection algorithm to replace S3FD for your specified target, i.e., RetinaFace~\cite{deng2019retinaface}, MTCNN~\cite{zhang2016joint}. 

\paragraph{Face Alignment} The second step is face alignment. 
After numerous experiments and failures, we realized that facial landmarks are the key to maintaining stability over time. We need to find an effective facial landmarks algorithm essential in producing an excellent successive footage shot and film.

DFL provides two canonical types of facial landmark extraction algorithms to solve this: (a) heatmap-based facial landmark algorithm 2DFAN~\cite{bulat2017far} (for faces with standard pose) and (b) PRNet~\cite{feng2018joint} with 3D face prior information (for faces with large Euler angle (yaw, pitch, roll), e.g., A face with a large yaw angle, means one side of the face is out of sight). 
After facial landmarks are retrieved, we also provide an optional function with a configurable time step to smooth facial landmarks of consecutive frames in a single shot to ensure stability further.
\begin{figure*}[h]
	\centering 
	[DF Structure] 
	\begin{subfigure}
	    \centering
	    
		\includegraphics[width=0.82\textwidth]{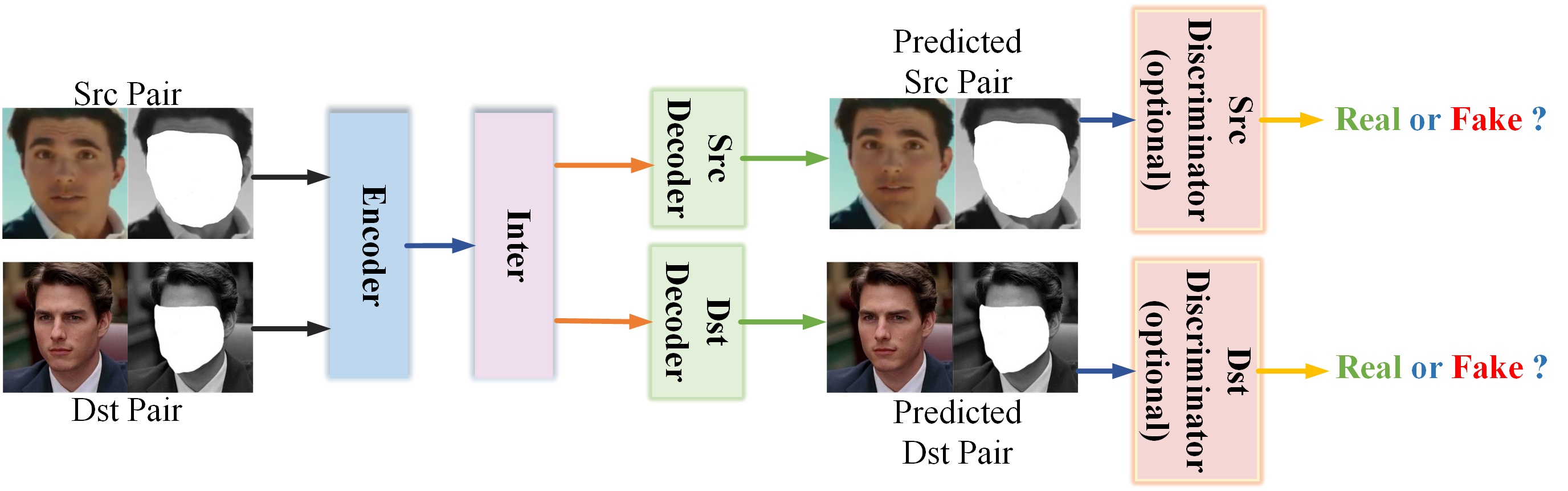}
	\end{subfigure}

    \begin{subfigure}
	    \centering
	    [LIAE Structure]
		\includegraphics[width=0.99\textwidth]{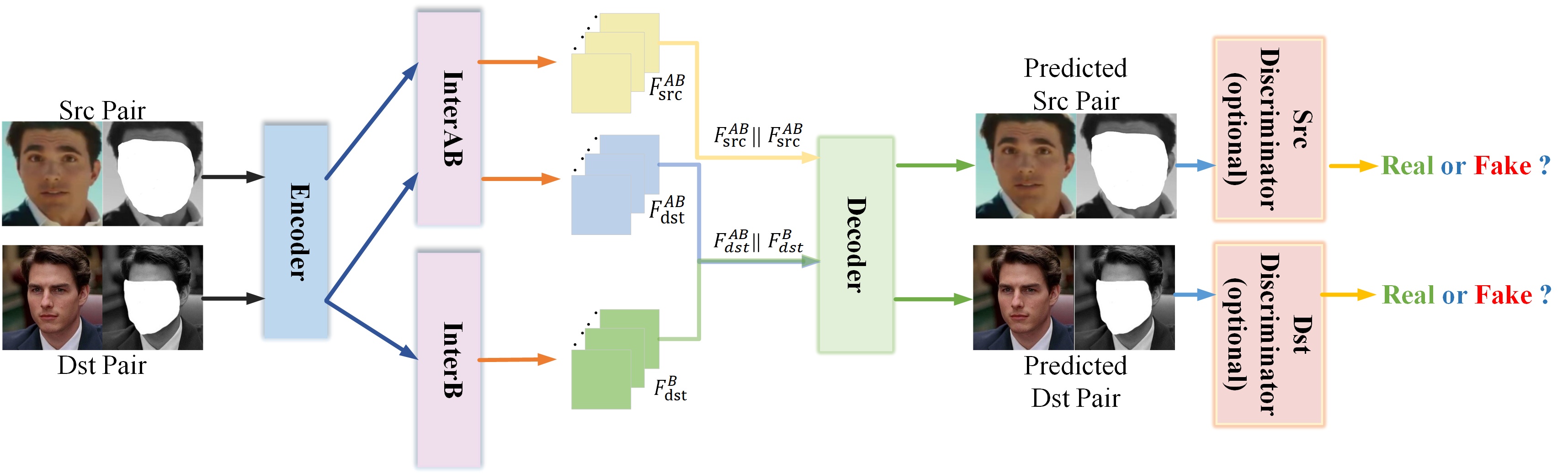}
	\end{subfigure}
	\caption{Overview of training phase in DeepFaceLab (DFL). DF structure and LIAE structure are both provided here for illustration, $\circ \Vert \circ$ represents the concatenation of latent vectors.}
	\label{fig:long3}
	\label{fig:onecol3}
\end{figure*}

Then we adopt a classical point pattern mapping and transformation method proposed by Umeyama~\cite{umeyama1991least} to calculate a similarity transformation matrix used for face alignment. 
As the method proposed by Umeyama et al. ~\cite{umeyama1991least} needs standard facial landmark templates in calculating similarity transformation matrix, DFL provides a canonical aligned facial landmark template. It is noteworthy that DFL could automatically predict the Euler angle by using the obtained facial landmarks.

\paragraph{Face Segmentation}\label{para:segmentation} After face alignment, a data folder with face of standard front/side-view ($\rm aligned$ $\rm src$ or $\rm aligned$ $\rm dst$) is obtained. We employ a fine-grained Face Segmentation network (TernausNet~\cite{iglovikov2018ternausnet}) on top of ($\rm aligned $ $\rm src$ or $\rm aligned $ $\rm dst$), through which a face with either hair, fingers, or glasses could be segmented exactly. It is optional but useful to remove irregular occlusions to keep the network in the training process robust to hands, glasses, and any other objects which may cover the face somehow.

However, since some state-of-the-art face segmentation models fail to generate fine-grained masks in some particular shots, the $\rm XSeg$ was introduced in DFL. $\rm XSeg$ allows everyone to train their model for the segmentation of a specific face set ($\rm aligned $ $\rm src$ or $\rm aligned $ $\rm dst$) through a few-shot learning paradigm (Figure~\ref{fig:xseg} is the schematic of $\rm XSeg$).

As the above workflow executed sequentially, everything DFL needs in the training phase is already prepared: cropped faces with their corresponding coordinates in its original images, facial landmarks, aligned faces, and pixel-wise segmentation masks from $\rm src$ (Since the extraction procedure of $\rm dst$ is the same with $\rm src$, there is no need to elaborate that in detail).

\subsection{Training}

\begin{figure*}[h]
	\begin{center}
		\includegraphics[width=0.8\linewidth]{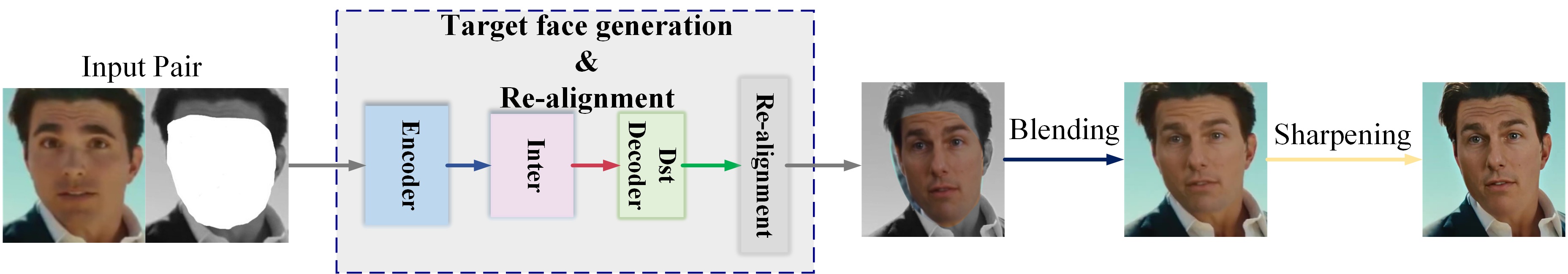}
	\end{center}
	\caption{Overview of conversion phase in DeepFaceLab(DFL).}
	\label{fig:long4}
	\label{fig:onecol4}
\end{figure*}

The training phase plays the most vital role in achieving photorealistic face-swapping results of DFL.

No need for facial expressions of $\rm aligned $ $\rm src$ and $\rm aligned $ $\rm dst$ being strictly matched, DFL aims to provide an efficient algorithm paradigm to solve this unpaired problem along with maintaining high fidelity and perceptual quality of the generated face. Motivated by the previous methods, we propose two structures, DF structure, and LIAE structure, to address this issue.

As shown in Figure~\ref{fig:long3}(a), DF structure consists of an $\rm Encoder$ as well as $\rm Inter$ with shared weights between $\rm src$ and $\rm dst$, two $\rm Decoder$s which belong to $\rm src$ and $\rm dst$ separately. The generalization of $\rm src$ and $\rm dst$ is achieved through the shared $\rm Encoder$ and $\rm Inter$, which solves the aforementioned unpaired problem easily. DF structure can finish the face-swapping task but cannot inherit enough information from $\rm dst$, such as lighting. 

To further enhance the problem of light consistency, we propose LIAE.
As depicted in Figure~\ref{fig:long3}(b), LIAE structure is a more complex structure with a shared-weight $\rm Encoder$, $\rm Decoder$ and two independent $\rm Inter$s. The main difference compared to the DF is that $\rm InterAB$ is used to generate both latent code of $\rm src$ and $\rm dst$ while $\rm InterB$ only output the latent code of $\rm dst$. Here, $F_{src}^{AB}$ denotes the latent code of $\rm src$ produced by $\rm InterAB$ and we generalize this representation to $F_{dst}^{AB}$, $F_{dst}^{B}$.
After getting all the latent codes from $\rm InterAB$ and $\rm InterB$, LIAE then concatenate these feature maps through channel: $F_{src}^{AB}||F_{src}^{AB}$ is obtained for a new latent code representation of $\rm src$ and $F_{dst}^{AB}||F_{dst}^{B}$ for $\rm dst$ as the same way. 

Then $F_{src}^{AB}||F_{src}^{AB}$ and $F_{dst}^{AB}||F_{dst}^{B}$ are put into the $\rm Decoder$ and we get the predicted $\rm src$ ($\rm dst$) alongside with their masks. The motivation of concatenating $F_{dst}^{B}$ with $F_{dst}^{AB}$ is to shift the direction of latent code in direction of the class ($\rm src$ or $\rm dst$) we need, through which $\rm InterAB$ obtained a compact and well-aligned representation of $\rm src$ and $\rm dst$ in the latent space.

Except for the structure of the model, some useful tricks are effective for improving the quality of the generated face: A weighted sum mask loss in general SSIM~\cite{wang2004image} can be added to make each part of the face carry different weights under the AE training architecture, for example, we add more weights to the eye area than the cheek, which aims to make the network concentrate on generating a face with vivid eyes.

As for losses, DFL uses a mixed loss (DSSIM (structural dissimilarity)~\cite{loza2006structural} + MSE) by default. The motivation for this combination is to get benefits from both: DSSIM generalizes human faces faster while MSE provides better clarity. This combination of losses serves to find a compromise between generalization and clarity.

Besides, we adopt a fancy true face mode $\rm TrueFace$, which serves for the generated face of better likeness to the $\rm dst$ in the conversion phase. For LIAE structure, we enforce $F_{src}^{AB}$ approaches $F_{dst}^{AB}$. And for DF structure
, counterparts turn out to be $F_{src}$ and $F_{dst}$. Two rarely used methods have been validated by DFL: Convolutional Aware Initialization~\cite{CAI2017} along with Learning Rate Dropout~\cite{LearningDropout2019}, which greatly enhanced the final quality of the fake face. 

%For further details about the network structure of DeepFaceLab, please refer to the ArXiv version$\footnote{https://arxiv.org/abs/2005.05535}$ for more details.

\subsection{Conversion}

% 2020.03.09 week11 星期一 face convert里面的图对应也要修改.

The conversion phase is the last but not least phase. Previous methods often ignore the importance of this phase. As depicted in Figure~\ref{fig:long4}, users can swap faces of $\rm src$ to $\rm dst$ and vice versa.

In the case of $\rm src2dst$, the first step of the proposed face-swapping scheme in the conversion phase is to transform the generated face alongside with its mask from $\rm dst$ $\rm Decoder$ to the original position of the target image in $\rm src$ due to the reversibility of Umeyama~\cite{umeyama1991least}.

% 2020.03.18 week12 星期三 rct, lct融合
The following piece is about blending, with the ambition for the realigned reenacted face to seamlessly fit with the target image along its outer contour.
For the sake of remaining consistent complexion, DFL provides five more color transfer algorithms (i.e., Reinhard color transfer: $\rm RCT$~\cite{reinhard2001color}, iterative distribution transfer: $\rm IDT$~\cite{pitie2007automated} and etc.) to approximate the color of the reenacted face to the target.
Any blending must account for different skin tones, face shapes, and illumination conditions, especially at the junctions between reenacted face with the delimited region and target face. DFL implemented this by Poisson blending~\cite{perez2003poisson}.

Finally, sharpening is indispensable. A pre-trained face super-resolution neural network was added to sharpen the blended face since it is noted that the generated faces in almost current state-of-the-art face-swapping works, more or less, are smoothed and lack minor details (i.e., moles, wrinkles).

%Finally, we will get a view of HD fake image. Generated face seamlessly into the designated part of target face and meanwhile adjust the skin tone of the generated face to the target face, then fit it back in the original picture according to its coordinates recorded in phase Extraction.

% 2020.03.16 week12 星期一
\section{Evaluation}

In this section, we compare the DeepFaceLab with several other commonly-used face-swapping frameworks and two state-of-the-art works. We find that DFL has competitive performance among them under identical experimental conditions.

\subsection{Qualitative results}
\label{section:Qualitative}

\begin{figure*}[htbp]
	\centering
	\subfigure[The comparison of DFL and representative open-source face-swapping projects.]{
		\begin{minipage}[t]{0.7\linewidth}
			\centering
			\includegraphics[width=0.94\linewidth]{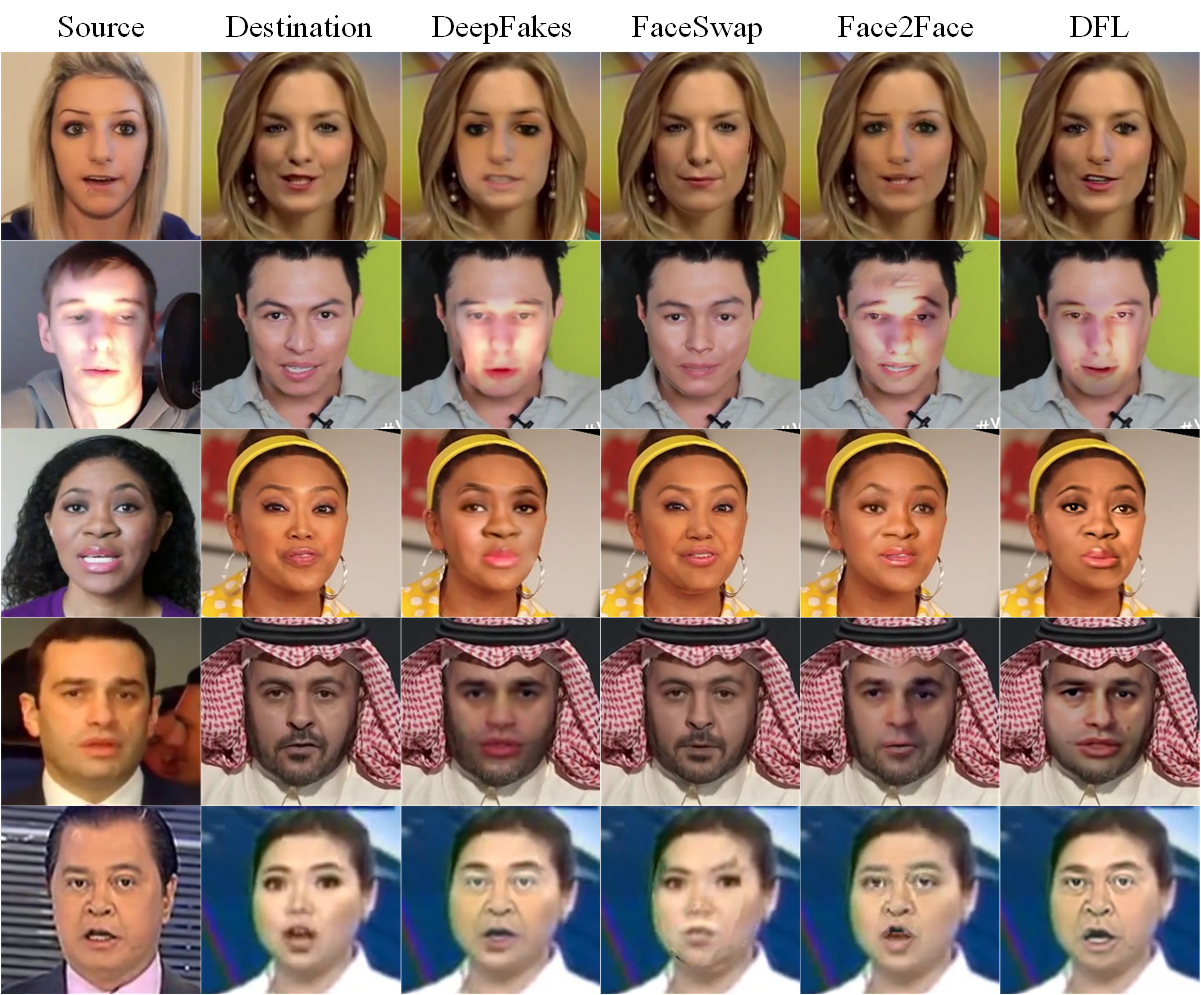}
		\end{minipage}
	}\\
	\subfigure[The comparison of DFL and the latest state-of-the-art face-swapping works.]{
		\begin{minipage}[t]{0.7\linewidth}
			\centering
			\includegraphics[width=0.94\linewidth]{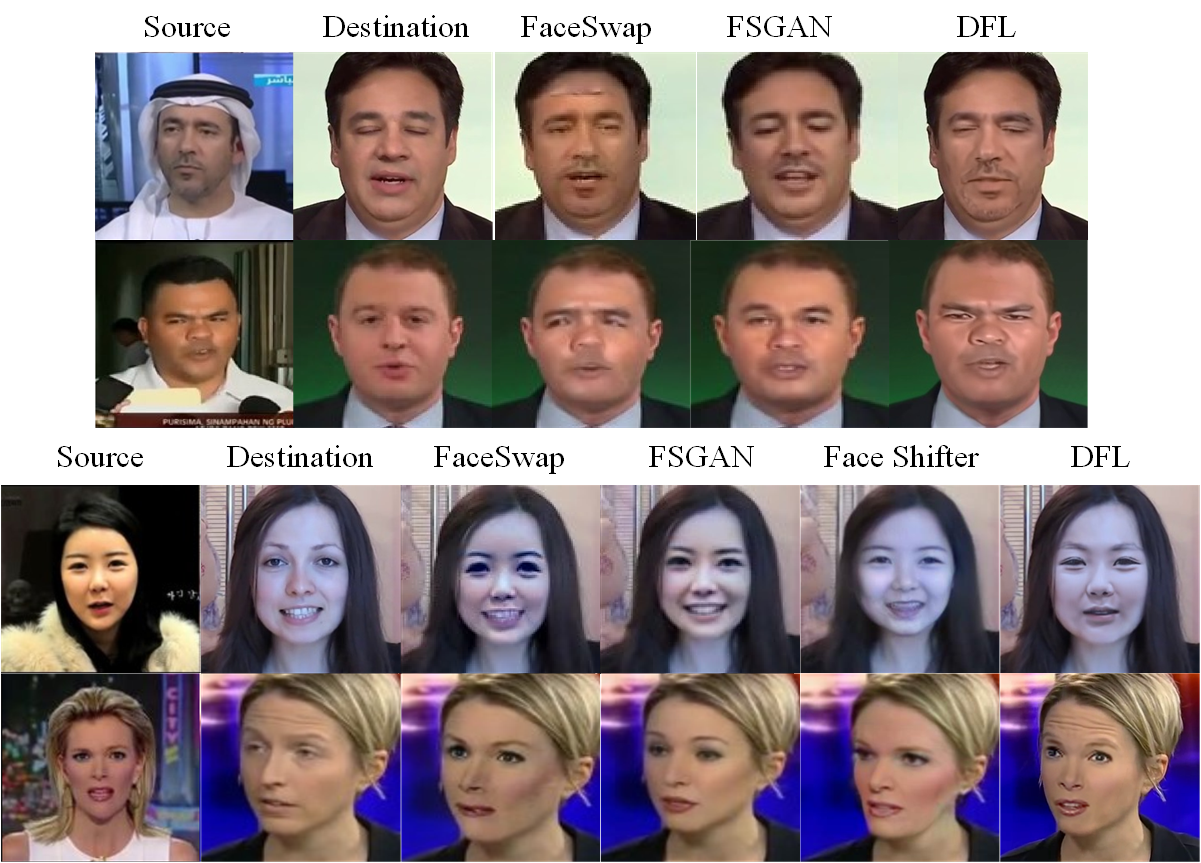}
		\end{minipage}%
	}
	\centering
	\caption{Qualitative face swapping results on FaceForensics++\cite{faceforensics} face images.}
	\label{fig:long62}
	\label{fig:onecol62}
\end{figure*}

Fig~\ref{fig:long62}(a) offers face-swapping results of representative open-source projects (DeepFakes~\cite{Deepfakes}, Nirkin et al.~\cite{nirkin2018_faceswap} and Face2Face~\cite{thies2016face2face}) taken from FaceForensics++ dataset~\cite{faceforensics}. Examples of different expressions, face shapes, and illuminations are selected in our experiment. It's clear from observing the video clips from FaceForensics++ that they are not only trained inadequately but chosen from models with low-resolution. To be fair in our comparison, $\rm Quick96$ mode is taken: a lightweight model that DF structure underneath, which outputs the $I_{output}$ of 96 $\times$ 96 resolutions (without $\rm GAN$ and $\rm TrueFace$). The average training time is restricted to 3 hours. We use Adam optimizer ($\rm lr$=0.00005, $\rm \beta_{1}$ = 0.5, $\rm \beta_{2}$ = 0.999) to optimize our model. All of our networks were trained on a single NVIDIA GeForce GTX 1080Ti GPU and an Intel Core i7-8700 CPU.

% 2020.03.19 week12 星期四 
% faceswap：https://github.com/ondyari/FaceForensics/tree/master/dataset/FaceSwapKowalski

% Deepfakes: https://github.com/ondyari/FaceForensics/tree/master/dataset/DeepFakes. 

% NT: Justus Thies, Michael Zollhofer, and Matthias Nießner. Deferred neural rendering: Image synthesis using neural textures. ACM Transactions on Graphics 2019 (TOG)

% face2face: Justus Thies, Michael Zollhofer,  Face2Face: Real-Time Face Capture and Reenactment of RGB Videos 016

\subsection{Quantitative results}

We compare our results with videos of FaceForensics++ in quantitative experiments. In practice, the naturalness and realness of the results of the face-swapping method are hard to describe with some specified quantitative indexes. However, pose and expression indeed embodies valuable insights of the face-swapping results. Besides, SSIM is used to compare the structure similarity as well as perceptual loss~\cite{johnson2016perceptual} is adopted to compare high-level differences, like content and style discrepancies, between the target subject and the swapped subject.

To measure the accuracy of the pose, we calculate the Euclidean distance between the Euler angles (extracted through FSA-Net~\cite{yang2019fsa}) of $I_{t}$ and $I_{output}$. Besides, the accuracy of the facial expression is measured through the Euclidean distance between the 2D landmarks (2DFAN~\cite{bulat2017far}). We use the default face verification method of DLIB~\cite{king2009dlib} for the comparison of identities.

To be statistically significant, we compute the mean and variance of those measurements on the 100 frames (uniform sampling over time) of the first 500 videos in FaceForensics++, averaging them across the videos. Here, DeepFakes~\cite{Deepfakes} and Nirkin et al.~\cite{nirkin2018_faceswap} are chosen as the baselines to compare. It should be noted that all the videos produced by DFL were followed by the same settings with ~\ref{section:Qualitative}.

% 2020.04.23 week17 星期四, 感知loss的文献.

\begin{table*}[th]
	\caption{Quantitative face swapping results on FaceForensics++~\cite{faceforensics} face images.}
	\label{tab:booktabs1}
	\centering
	\setlength{\tabcolsep}{1.5mm}{
		\begin{tabular}{llllll}
			\toprule
			Method & SSIM $\uparrow$ & perceptual loss $\downarrow$ & verification $\downarrow$  & landmarks $\downarrow$ & pose $\downarrow$ \\
			\midrule
			DeepFakes & 0.71 $\pm$ 0.07 & 0.41 $\pm$ 0.05 & 0.69 $\pm$ 0.04 & 1.15 $\pm$ 1.10 & 4.75 $\pm$ 1.73    \\
			Nirkin et al. & 0.65 $\pm$ 0.08 & 0.50 $\pm$ 0.08 & 0.66 $\pm$ 0.05 & \bf{0.35} $\pm$ \bf{0.18} & 6.01 $\pm$ 3.21    \\
			DFL(ours) & \bf{0.73} $\pm$ \bf{0.07} & \bf{0.39} $\pm$ \bf{0.04} & \bf{0.61} $\pm$ \bf{0.04} & 0.73 $\pm$ 0.36 & \bf{1.12} $\pm$ \bf{1.07}    \\
			\bottomrule
	\end{tabular}}
\end{table*}

From the indicators listed in Table~\ref{tab:booktabs1}, DFL is more adept at retaining pose and expression than baselines. Besides, with the empowerment of super-resolution in the conversion phase, DFL often produces $I_{output}$ with vivid eyes and sharp teeth, but this phenomenon couldn't be reflected clearly in the SSIM-like score for they only take a small part of the whole face.

\subsection{Ablation study}

To compare the visual effects of different model choices, GAN settings and etc., we perform several ablation tests. The ablation study is conducted on top of three essential parts: network structure, training paradigm, and latent space constraint.

\begin{figure}[h]
	\begin{center}
		\includegraphics[width=0.95\linewidth]{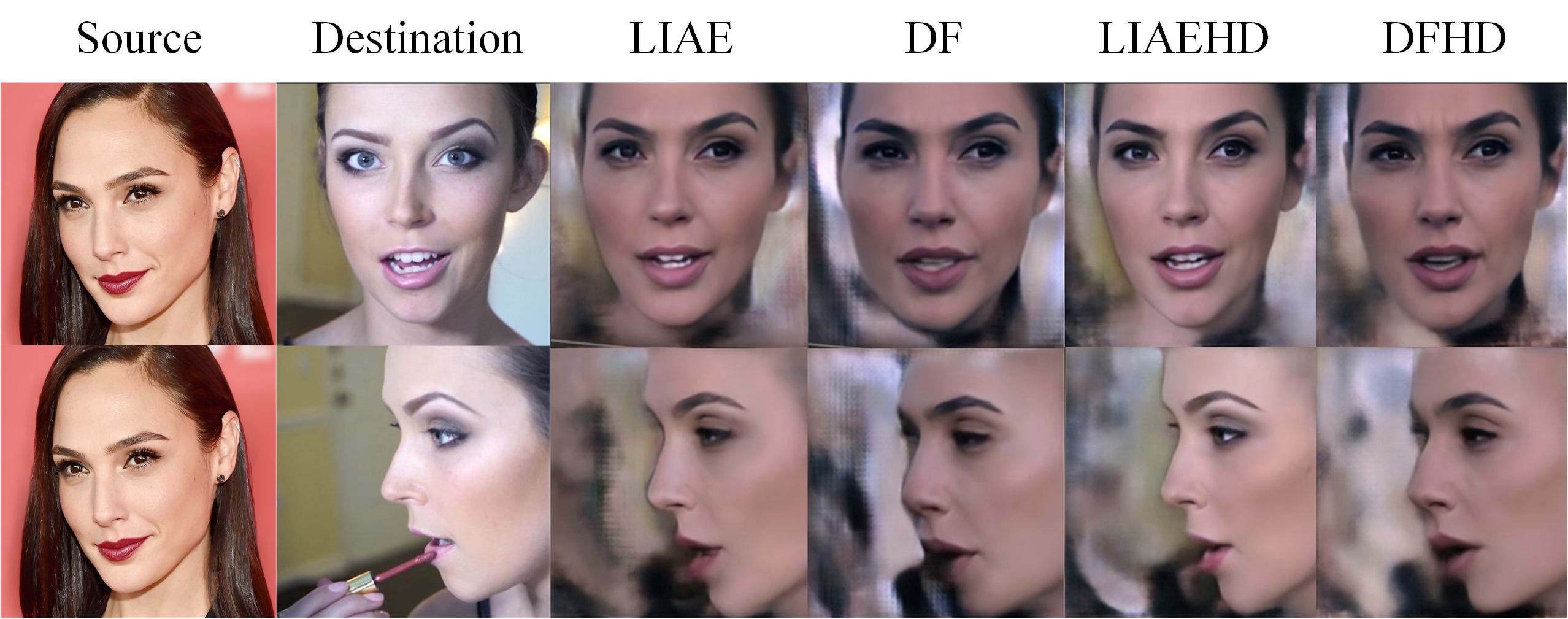}
	\end{center}
	\caption{Ablation experiments of different model structures (with $\rm GAN$ and $\rm TrueFace$). (Here, we provide training previews instead of the converted faces, which aims to make a fair comparison in model architectures of DFL meanwhile avoid the impact of post-processing from the conversion phase.)}
	\label{fig:long51}
	\label{fig:onecol51}
\end{figure}

\begin{figure}[htpb]
	\begin{center}
		\includegraphics[width=0.95\linewidth]{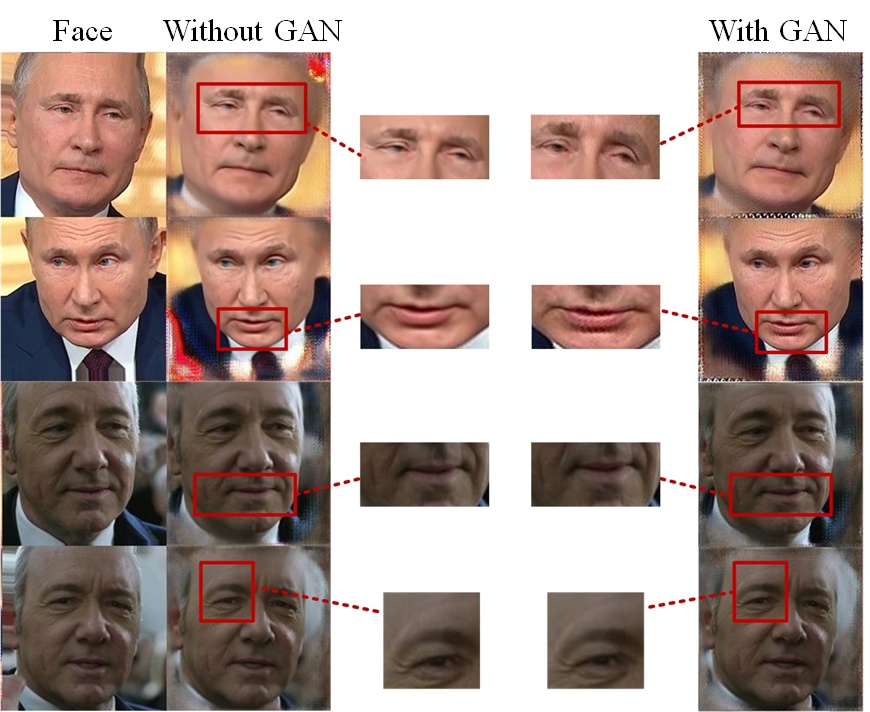}
	\end{center}
	\caption{Ablation experiments of different training paradigms: non-GAN-based and GAN-based (The image on the left is the original face, a reconstruction image produced by a model that trained without GAN listed to its right, far right is produced by a model trained with GAN). Obviously, GAN enforces the model to become more sensible in capturing the sharp details, i.e., wrinkles and moles. Meanwhile, significantly reduce the vagueness compared to the model without the empower of GAN.}
	\label{fig:long52}
	\label{fig:onecol52}
\end{figure}

Aside from DF structure and LIAE structure, we also enhance them to DFHD and LIAEHD by adding more feature extraction layers and residual blocks than the original version, which enriches the model structures for comparison. Related details are demonstrated in the supplemental material. The qualitative results of different model structures can be seen in Figure~\ref{fig:long51}, and the qualitative results of different training paradigms are depicted in Figure~\ref{fig:long52}. As shown in Figure~\ref{fig:long51}, we can see that LIAE can inherit a well-shaped face shape from $\rm dst$ and generate more advanced results than DF, which solves the unmatched face shape problem gracefully. Moreover, to probe whether the introduction of GAN works in DFL, we compare GAN-based with non-GAN-based in Figure~\ref{fig:long52}, it is apparent that the details of the face are more realistic and lumpy than non-GAN generated.

Quantitative ablation results are reported in Table~\ref{tab:booktabs2}. The experiment settings of the training are almost the same with ~\ref{section:Qualitative} except for the structure of the model.

Verification results from Table~\ref{tab:booktabs2} show that source identities are preserved across networks with the same structure. With more shortcut connections added to the model (i.e., DF to DFHD, LIAE to LIAEHD), scores of landmarks and pose decrease without $\rm GAN$. Meanwhile, the generated results could have a better chance to get rid of the influence of the source face.

Also, we found that $\rm TrueFace$ is effectively relieved the instability of $\rm GAN$, through which a more photo-realistic result without much degradation is then achieved. Besides, SSIM progressively increases with more shortcut connections, $\rm TrueFace$ and $\rm GAN$ also do good to it in varying degrees.

% 2020.03.14 手动干预控制合成的情况.

\begin{table*}[htbp]
	\caption{Quantitative ablation results on FaceForensics++ ~\cite{faceforensics} face images.}
	\label{tab:booktabs2}
	\centering
	\setlength{\tabcolsep}{1.5mm}{
		\begin{tabular}{lllll}
			\toprule
			Method & SSIM $\uparrow$ & verification $\downarrow$  & landmarks $\downarrow$ & pose $\downarrow$ \\
			\midrule
			DF & 0.73 $\pm$ 0.07  & 0.61 $\pm$ 0.04 & 0.73 $\pm$ 0.36 & 1.12 $\pm$ 1.07     \\
			DFHD & 0.75 $\pm$ 0.09  & 0.61 $\pm$ 0.04 & 0.71 $\pm$ 0.37 & 1.06 $\pm$ 0.97     \\
			DFHD ($\rm GAN$) & 0.72 $\pm$ 0.11  & 0.61 $\pm$ 0.04 & 0.79 $\pm$ 0.40 & 1.33 $\pm$ 1.21     \\
			DFHD ($\rm GAN + TrueFace$) & 0.77 $\pm$ 0.06  & 0.61 $\pm$ 0.04 & 0.70 $\pm$ 0.35 & 0.99 $\pm$ 1.02    \\
			LIAE & 0.76 $\pm$ 0.06  & \bf{0.58} $\pm$ \bf{0.03} & 0.66 $\pm$ 0.32 & 0.91 $\pm$ 0.86   \\
			LIAEHD & 0.78 $\pm$ 0.06  & 0.58 $\pm$ 0.03 & \bf{0.65} $\pm$ \bf{0.32} & 0.90 $\pm$ 0.88   \\
			LIAEHD ($\rm GAN$) & 0.79 $\pm$ 0.05  & 0.58 $\pm$ 0.03 & 0.69 $\pm$ 0.34 & 1.00 $\pm$ 0.97   \\
			LIAEHD ($\rm GAN + TrueFace$) & \bf{0.80} $\pm$ \bf{0.04}  & 0.58 $\pm$ 0.03 & 0.65 $\pm$ 0.33 & \bf{0.83} $\pm$ \bf{0.81}   \\
			\bottomrule
	\end{tabular}}
\end{table*}
\section{Discussion}
\subsection{Integrity}
Previous methods often lack integrity. As mentioned in Sec.\ref{section:pipeline}, DFL consists of three main phases, extraction, training, and conversion. Each phase plays a different role and has various kinds of alternative techniques. Thanks to the long development progress, DFL has become the most mature face-swapping system in the world. For example, we provide several kinds of face segmentation methods. It is noteworthy that DFL is not a simple combination of current state-of-the-art methods. Instead, most efficient tools are developed by ourselves according to users' requirements.

\begin{figure}[htpb]
	\begin{center}
		\includegraphics[width=0.95\linewidth]{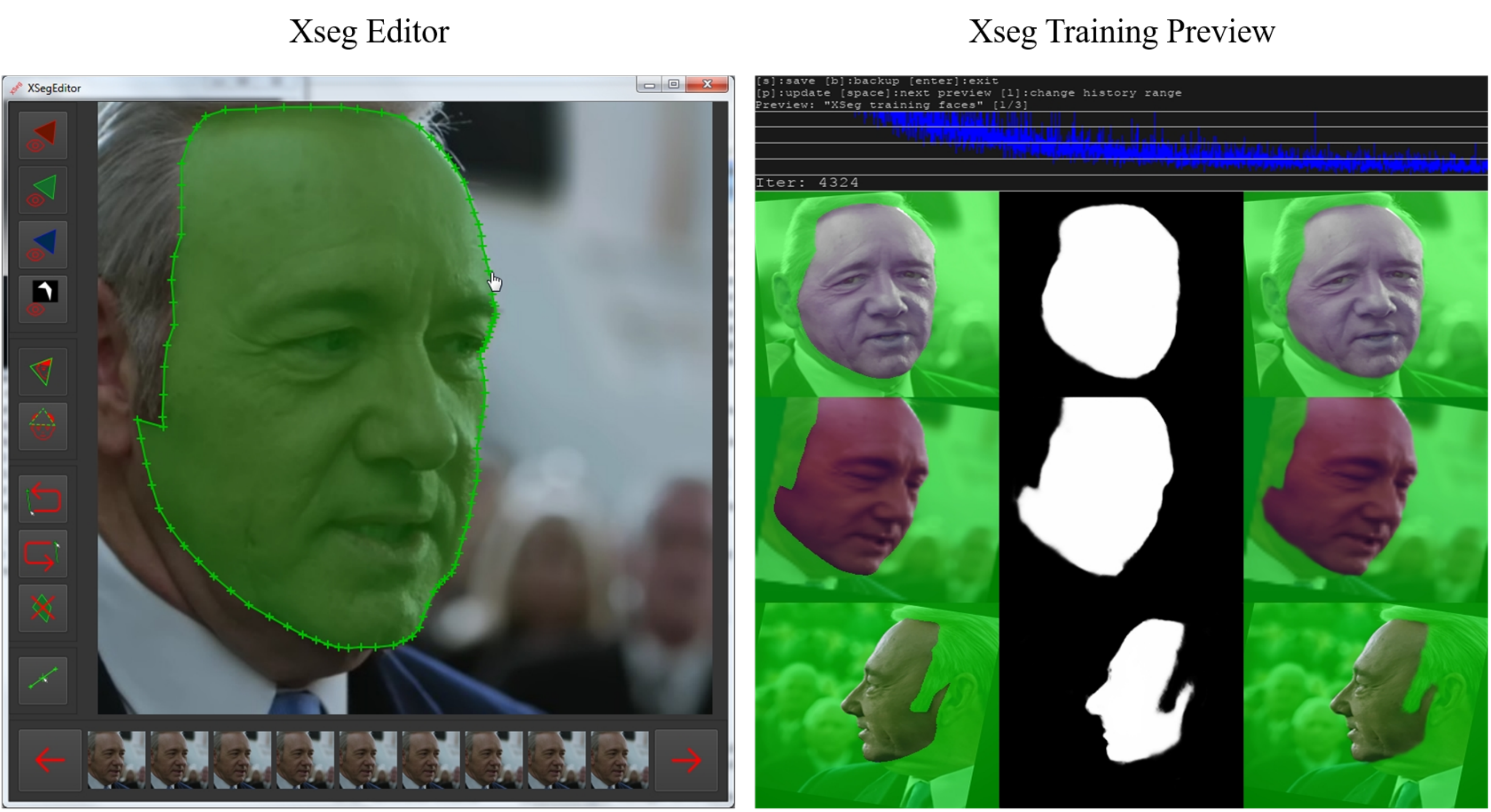}
	\end{center}
	\caption{The preview of $\rm XSeg$. Users can label the masks they want by $\rm XSeg Editor$. With the help of $\rm XSeg$, users can use it to eliminate the occlusion of hands, glasses, and any other objects which may cover the face somehow and control specific areas for swapping.}
	\label{fig:xseg}
\end{figure}

Considering the complex requirements for face segmentation, DFL provides an automatic algorithm, TernausNet\cite{iglovikov2018ternausnet} as default. As mentioned in Section \ref{para:segmentation}, TernausNet can remove irregular occlusions efficiently. However, this model may fail to generate fine-grained masks in some particular shots. So we develop a high-efficiency face segmentation tool, $\rm XSeg$, which allows everyone to customize to suit specific needs by few-shot learning. As shown in Figure \ref{fig:xseg}, users can create the mask label and train a customized segmentation model. With the help of $\rm XSeg$, users can define the swapping mask by themselves and solve almost all problems in the extraction and conversion phase.
\subsection{Potential}
%The application scenarios of previous face-swapping methods greatly limit the horizon of previous researchers. 
Previous methods usually focus on synthesizing high-quality results by feeding two videos or hundreds of images. However, making good face-swapping results in this manner is not reasonable at all. DFL supports huge-scale datasets, up to $\sim$ 100k images. With the help of enormous data, final swapped results can achieve significant quality improvements.

Besides, previous methods always lack potentials. Unlike previous methods, DFL can provide face-swapping and support lip manipulation, head replacement, do-age and etc. These new functions can be realized by simply finetuning our framework. We also encourage users to use video editing tools, such as DaVinci Resolve and After Effect, further enhance the visual quality of the final video.

\subsection{Broader Impact}
%Any Face swapping algorithm runs the risk of producing biased or unsuitable content---our work is certainly not an exception.

Since the attention deepfake-related productions received grew exponentially, DFL, the most commonly used deepfake generation tool for VFX artists, has played an irreplaceable role. The emergence of DFL certainly adds to the entertainment to the world meanwhile of high economic value in the post-production industry when it comes to replacing the stunt actor with pop stars. 

Because DFL could produce the cinema-quality face-swapping result, researchers who engage in forgery detection areas may be motivated to design robust classifiers for high-quality forgery video clips or images.

%Inspired by some distinguished researchers of this area: "Suppressing the publication of such methods would not stop their development, but rather make them only available to a limited number of experts and potentially blindside policy makers if it goes without any limits", we believe that it's our responsibility to publish DeepFaceLab to the academia community formally.
\section{Conclusions}
The rapidly evolving DeepFaceLab has become a popular face-swapping tool in the deep learning practitioner community by freeing people from laborious, complicated data processing, trivial detailed work in training, and conversion. As more and more people participate in the development of DeepFaceLab, deepfake entertainment has been trending in social media. In the future, we want to dig deeper into the entertainment-related AI framework while pushing the forgery detection field forward.
%While keeping tight with the latest trends and advances in computer vision, in the future, we plan to keep improving the speed and scalability of DeepFaceLab.

\clearpage
{\small
\bibliographystyle{ieee_fullname}
\bibliography{egbib}
}

\end{document}